\documentclass{svproc}

\pdfoutput=1

\usepackage{amsmath,amssymb,amsfonts}
\usepackage{algorithmic}
\usepackage{graphicx}
\usepackage{url}
\usepackage{hyperref}
\usepackage{xcolor}
\usepackage{textcomp}
\usepackage{booktabs}

\usepackage{microtype}
\usepackage{url}
\usepackage{lmodern}
\usepackage[T1]{fontenc}
\usepackage[utf8]{inputenc}
\usepackage{graphicx}
\usepackage{amsmath}
\usepackage{xcolor}
\usepackage{colortbl}
\usepackage{tikz}
\usepackage{booktabs}
\usepackage{multirow}
\usepackage{authblk}
\usepackage{hyperref}
\usepackage{float}

\makeatletter
\renewcommand{\inst}[1]{\textsuperscript{#1}}
\makeatother
\begin{document}
\mainmatter  


\title{LLM-SEM: A Sentiment-Based Student Engagement Metric Using LLMS for E-Learning Platforms}

\titlerunning{LLM-SEM: Measuring Student Engagement in E-Learning Platforms}

\author{
    Ali Hamdi \inst{1} \and
    Ahmed Abdelmoneim Mazrou\inst{1} \and
    Mohamed Shaltout\inst{2}
}
\institute{
    \inst{1} AiTech AU, Preston, Australia \\
    \email{ali@aitech.net.au}, 
    \email{ahmed.abdelmoneim@aitech.net.au}\\
    \inst{2} Mawhiba, Riyadh, KSA, \email{shaltot@icloud.com}
}

\maketitle  
\begin{sloppypar}

\begin{abstract}
Current methods for analyzing student engagement in e-learning platforms, including automated systems, often struggle with challenges such as handling fuzzy sentiment in text comments and relying on limited metadata. Traditional approaches, such as surveys and questionnaires, also face issues like small sample sizes and scalability. In this paper, we introduce LLM-SEM (Language Model-Based Student Engagement Metric), a novel approach that leverages video metadata and sentiment analysis of student comments to measure engagement. By utilizing recent Large Language Models (LLMs), we generate high-quality sentiment predictions to mitigate text fuzziness and normalize key features such as views and likes. Our holistic method combines comprehensive metadata with sentiment polarity scores to gauge engagement at both the course and lesson levels. Extensive experiments were conducted to evaluate various LLM models, demonstrating the effectiveness of LLM-SEM in providing a scalable and accurate measure of student engagement. We fine-tuned TXLM-RoBERTa using human-annotated sentiment datasets to enhance prediction accuracy and utilized LLama 3B, and Gemma 9B from Ollama.
\end{abstract}

\section{Introduction}
\label{sec:introduction}

The proliferation of e-learning platforms has transformed the way students engage with educational content \cite{rothwell2024revolutionizing}. Platforms such as YouTube host numerous educational channels that cater to diverse learning needs \cite{maynard2021succeed}. However, assessing student engagement with these resources remains a significant challenge for educators and content creators \cite{heilporn2021examination}. Traditional methods, such as surveys and questionnaires, often suffer from small sample sizes, lack of scalability, and difficulty in capturing the opinions of a broader audience~\cite{ituma2011evaluation}. Automated methods, while an improvement, face their own challenges, including dealing with fuzzy sentiment in text comments and relying on limited metadata, which hampers their ability to provide a comprehensive assessment of engagement \cite{saka2023gpt}.

Sentiment analysis has been advancing applications across several domains \cite{hamad2022attention,badaro2019survey,baly2017omam,hamad2022steducov,hamad2024asem,hamad2022empathy}. Current sentiment analysis methods for student engagement, while useful, are often hindered by the ambiguity of user comments, leading to misinterpretations of student feedback. Furthermore, traditional metrics heavily rely on small-scale metadata or subjective self-reports, which do not scale well for large e-learning platforms. There is a need for a scalable, data-driven metric that can integrate both qualitative sentiment data and quantitative metadata to assess student engagement more accurately.

To address these limitations, we propose \textbf{LLM-SEM} (Language Model-Based Student Engagement Metric), a novel quantitative method to gauge student engagement on e-learning platforms. By leveraging recent advances in Large Language Models (LLMs), LLM-SEM combines course and lesson video metadata with sentiment analysis of user comments, providing a more accurate and scalable solution. This approach overcomes the constraints of traditional methods by incorporating comprehensive metadata and high-quality sentiment predictions, allowing for a holistic evaluation of student engagement at both the course and lesson levels.

Our research introduces an LLM-driven metric that addresses the challenges faced by current methods and provides a scalable solution to engagement analysis. We fine-tuned the RoBERTa model on human-annotated sentiment datasets to generate high-quality sentiment predictions that enhance the analysis of student feedback. Other state-of-the-art LLMs, such as TXLM-RoBERTa~\cite{liu2019roberta}, LLama 3B~\cite{touvron2023llama}, and Gemma 9B~\cite{team2024gemma}, were evaluated without fine-tuning. Additionally, we introduce a methodology that normalizes key metadata features such as views and likes, integrating these with sentiment polarity scores to form a comprehensive engagement metric.

This paper is organized as follows: Section~\ref{sec:related_work} discusses related work on student engagement and sentiment analysis. Section~\ref{sec:methodology} outlines our proposed methodology and the LLM-SEM model. Section~\ref{sec:Results_Discussion} presents the experimental results and discussion of the findings. Finally, Section~\ref{sec:conclusion} concludes the paper.

\section{Related Work}
\label{sec:related_work}

Student engagement and feedback are crucial indicators of the effectiveness of educational content \cite{ogunyemi2022indicators}. Traditionally, course evaluation methods have relied on qualitative approaches, such as surveys and questionnaires, to gather feedback on student engagement and satisfaction~\cite{mandouit2018using}. These methods, while valuable, face limitations in scalability, representativeness, and timeliness. With the growing volume of online educational content, there has been a shift towards automated and data-driven methods for assessing student engagement, including sentiment analysis and the use of metadata \cite{friedman2024using}.

\setlength{\tabcolsep}{0.8pt}

Sentiment analysis has been widely applied across various domains to gauge public opinion, including politics, marketing, and consumer reviews~\cite{hamad2024asem,drus2019sentiment,badaro2019survey,baly2017omam,hamdi2018clasenti,hamdi2016review}. In the context of education, sentiment analysis has emerged as a promising tool for evaluating student satisfaction and engagement. Previous studies have demonstrated the potential of sentiment analysis in predicting student satisfaction by analyzing reviews and comments from various online learning platforms.

Recent advancements in Large Language Models (LLMs) have revolutionized sentiment analysis, enabling more accurate and nuanced predictions of sentiment in textual data~\cite{zhang2023enhancing}. LLMs such as BERT \cite{kenton2019bert} and its variations, including AraBERT and RoBERTa, have shown significant improvements in understanding the sentiment of complex and context-dependent text, addressing the limitations of earlier sentiment analysis methods. These models have been fine-tuned on a variety of sentiment datasets to enhance their performance in domain-specific applications. Leveraging LLMs for sentiment analysis in educational contexts can help overcome the challenges of fuzzy and ambiguous student comments, providing more accurate insights into student satisfaction~\cite{antoun2020arabert}.

Combining sentiment polarity scores with video metadata is another area of growing interest in the educational domain. Metadata such as views, likes, and the duration of student interaction with course materials can provide valuable insights into engagement levels~\cite{ginda2019visualizing}. Some studies have proposed methods that incorporate both sentiment analysis and metadata to compute engagement scores, particularly in non-educational settings like social media and entertainment platforms~\cite{pereira2018broad}. However, applying these techniques to educational platforms, especially on YouTube, is still in its early stages. By integrating comprehensive metadata with sentiment analysis, we can achieve a more holistic and scalable approach to measuring student engagement.

In this paper, we build on these developments by proposing a novel metric, \textbf{LLM-SEM}, which combines sentiment analysis using LLMs with course and lesson metadata to compute student engagement scores. Our approach addresses the limitations of current methods by incorporating both the textual sentiment and the metadata features to gauge student engagement more accurately and at scale.

\section{Methodology}
\label{sec:methodology}

\begin{figure}[H]
\centering
\includegraphics[width=.9\textwidth]{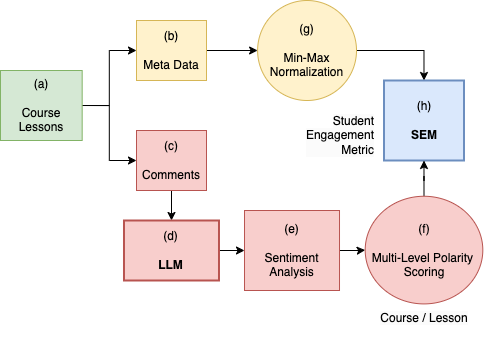}
\caption{A workflow illustrating the process of calculating the Student Engagement Metric (SEM) using course lessons, metadata, comments, and sentiment analysis. Starting from (a) the course lessons, (b) metadata is extracted and normalized using Min-Max normalization (g) to contribute to the final SEM (h). Simultaneously, (c) comments are analyzed by a Large Language Model (LLM) (d) for sentiment analysis (e), leading to (f) multi-level polarity scoring. The scoring is aggregated with the normalized metadata to compute the final student engagement metric for each course or lesson.}
\label{fig:architecture}
\end{figure}

Our methodology involves several stages, each representing a distinct step in the process of calculating the Student Engagement Metric (SEM). These stages include data collection, metadata extraction, sentiment analysis, polarity scoring, feature space normalization, and finally, the calculation of a comprehensive engagement metric by integrating both sentiment and metadata. A high-level workflow is shown in Fig.~\ref{fig:architecture}, which demonstrates how course lessons, metadata, and comments are processed to compute the final SEM.

\subsection{Data Collection and Manipulation}
\begin{figure}[H]
\centering
\includegraphics[width=\textwidth]{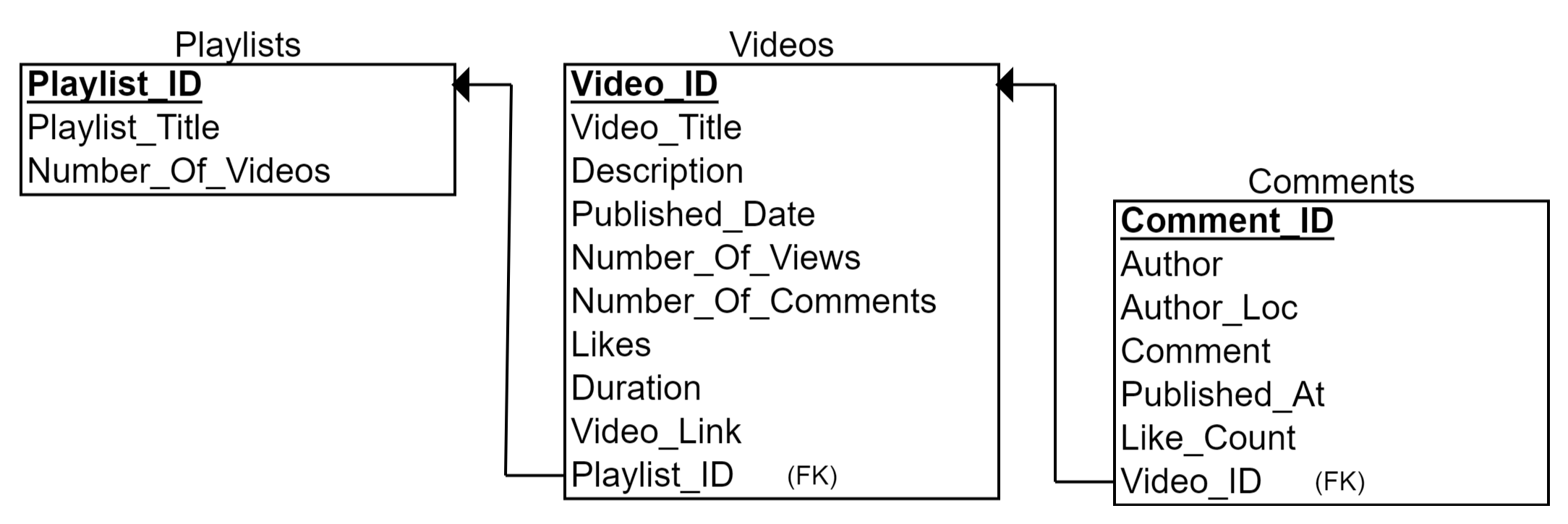}
\caption{Schema of the data collection structure showing the relationships between Playlists, Videos, and Comments. Each playlist contains multiple videos, identified by a foreign key \texttt{Playlist\_ID}. Videos are linked to comments through the \texttt{Video\_ID} foreign key, and each video has associated metadata such as views, likes, and duration. This relational structure allows efficient organization and processing of data for analysis.}
\label{fig:data_schema}
\end{figure}

Our data collection process focuses on two levels of granularity: course and lesson. Each course is represented by a playlist, and each lesson corresponds to individual videos within that playlist. We scraped data from each channel, organizing it into three main categories: playlists, videos, and comments. The structure of these categories and their relationships are illustrated in Fig.~\ref{fig:data_schema}.

The collected data is organized into two primary types:
\begin{itemize}
    \item \textbf{Metadata:} This includes video-specific details such as the number of views, likes, video length, and publication date. Metadata features play a crucial role in measuring the quantitative aspects of engagement. Key metadata extracted for each video includes:
        \begin{itemize}
            \item \textbf{Views:} The total number of views each video has received.
            \item \textbf{Likes:} The number of likes on each video, which can indicate user appreciation of the content.
        \end{itemize}
    \item \textbf{Textual Data (Comments):} This includes user comments associated with each video, providing a rich source of qualitative insights into how students perceive the course or lesson content. These comments are used for sentiment analysis to assess user sentiment and engagement. 
\end{itemize}

This structured collection of metadata and comments allows us to assess engagement on both quantitative (views, likes) and qualitative (comments, sentiment) levels. Fig.~\ref{fig:data_schema} outlines the schema of this data collection process, illustrating the relationships between Playlists, Videos, and Comments, and how they are organized for further analysis.

\subsection{LLM for Sentiment Analysis}
To generate sentiment scores from user comments, we employed several state-of-the-art Large Language Models (LLMs). These models help to parse and interpret the sentiment behind each comment, classifying it as positive, negative, or neutral. The following LLMs were tested in our experiments:
\begin{itemize}
    \item \textbf{Twitter-XLM-RoBERTa-Base for Sentiment Analysis:} A multilingual model fine-tuned on sentiment analysis for tweets across various languages~\cite{barbieri2020tweeteval}.
    \item \textbf{LLama 3B from Ollama:} A 3-billion parameter model designed for multilingual tasks, capable of performing sentiment analysis with high accuracy.
    \item \textbf{Gemma 9B from Ollama:} A larger 9-billion parameter multilingual model that offers improved performance in sentiment classification tasks.
\end{itemize}

\textbf{Human-Annotated Sentiment Dataset:} For fine-tuning and evaluation, we used a human-annotated sentiment dataset sourced from the [Arabic Sentiment Corpora GitHub Repository](https://github.com/ali7amdi/Arabic-Sentiment-Corpora). This dataset contains a diverse set of text samples with manually labeled sentiment categories (positive, negative, and neutral), making it suitable for training and evaluating sentiment models in an educational context. Since the metadata collected for this study was primarily in Arabic, it was crucial to test the models on an Arabic-specific human-annotated sentiment dataset to ensure high-quality sentiment analysis for the student engagement evaluation.

For future work, we plan to expand the sentiment analysis by exploring human-annotated sentiment datasets in other languages to improve the versatility and robustness of the model for diverse educational platforms.

\subsection{Sentiment Analysis}
Once the comments are processed by the LLM, each comment is assigned a sentiment label (positive, negative, or neutral) along with a confidence score, which ranges from -1 to 1. These scores are then used to compute a polarity score for each lesson in each course. This stage is crucial for capturing the qualitative aspect of student engagement, as sentiment provides insight into how users feel about the course or lesson content.

\subsection{Multi-Level Polarity Scoring (f)}
For each video \( v \), we compute a polarity score \( P_v \) based on the sentiment of user comments. This score ranges from \( -1 \) (entirely negative) to \( 1 \) (entirely positive), with \( 0 \) indicating neutral sentiment. The polarity score is calculated by assigning weighted sentiment scores to each comment based on the model’s confidence in the sentiment classification.

For a comment \( c \) with sentiment \( s_c \) and confidence score \( \text{score}_c \), the weighted score \( w_c \) is determined as:

\[
w_c =
\begin{cases} 
\text{score}_c, & \text{if } s_c = \text{positive} \\
-\text{score}_c, & \text{if } s_c = \text{negative} \\
0, & \text{if } s_c = \text{neutral}
\end{cases}
\]

The polarity score \( P_v \) for video \( v \) is then calculated by averaging the weighted scores of all comments:

\[
P_v = \frac{\sum_{c \in C_v} w_c}{N_v}
\]

where \( C_v \) is the set of comments on video \( v \), and \( N_v \) is the total number of comments. By dividing the sum of weighted scores by the total number of comments, we normalize the polarity score to ensure it remains within the range of \( -1 \) to \( 1 \), regardless of the number of comments per video. This normalization allows for fair comparison between videos with different levels of engagement, ensuring that videos with fewer comments are not unfairly penalized or advantaged when calculating their engagement score.

Beyond individual videos, we extend our analysis to compute a playlist polarity score \( P_p \) for each playlist \( p \). This score represents the average sentiment across all videos within a playlist, providing insights into the overall reception of the playlist’s content.

For each playlist \( p \), which consists of a set of videos \( V_p \), the playlist polarity score \( P_p \) is calculated as the mean of the polarity scores of its constituent videos:

\[
P_p = \frac{\sum_{v \in V_p} P_v}{N_p}
\]

where:
\begin{itemize}
    \item \( P_v \) is the polarity score of video \( v \) within playlist \( p \), as previously defined.
    \item \( N_p = |V_p| \) is the total number of videos in playlist \( p \).
\end{itemize}

This aggregation method ensures that each video’s polarity contributes equally to the overall playlist polarity, regardless of the number of comments on individual videos. By averaging the video polarities, we normalize for varying video lengths and comment volumes, allowing for a fair comparison between playlists.

\subsection{Feature Space Normalization}
To ensure that metadata (such as views and likes) and sentiment polarity scores are comparable across videos with varying levels of engagement, we apply min-max normalization to these features. The normalized values are scaled between 0 and 1, ensuring that all features contribute proportionally to the final engagement metric.

Let:
\begin{itemize}
    \item \( V_v \) be the number of views for video \( v \),
    \item \( L_v \) be the number of likes for video \( v \),
    \item \( V_{\min} \) and \( V_{\max} \) represent the minimum and maximum views across all videos,
    \item \( L_{\min} \) and \( L_{\max} \) represent the minimum and maximum likes across all videos,
    \item \( P_v \) be the sentiment polarity score for video \( v \).
\end{itemize}

The normalized views for a video \( v \), denoted as \( NV_v \), are calculated as:

\[
NV_v = \frac{V_v - V_{\min}}{V_{\max} - V_{\min}}
\]

Similarly, the normalized likes for a video \( v \), denoted as \( NL_v \), are calculated as:

\[
NL_v = \frac{L_v - L_{\min}}{L_{\max} - L_{\min}}
\]

\subsection{LLM-SEM: A Novel Engagement Metric}
Our proposed engagement metric, LLM-SEM (Language Model-Based Student Engagement Metric), integrates the normalized video metadata (views and likes) with the sentiment polarity scores to compute an overall engagement score, denoted as \( E_v \), for each video. This metric provides a comprehensive measure of student engagement by combining both qualitative (sentiment) and quantitative (views and likes) data.

The overall engagement score is calculated by summing the normalized views (\( NV_v \)), normalized likes (\( NL_v \)), and the sentiment polarity score (\( P_v \)) as:

\[
E_v = NV_v + NL_v + P_v
\]

This formula combines metadata and sentiment data into a single score, enabling fair comparisons across videos and courses with varying levels of interaction.

\subsubsection{Range and Interpretation of the Engagement Metric \( E_v \)}
Since each component is normalized, \( E_v \) will fall within the range of -1 to 3:
\begin{itemize}
    \item \textbf{Maximum Value (3):} Represents the highest engagement, achieved when views and likes are maximized (both equal to 1) and sentiment is entirely positive (\( P_v = 1 \)).
    \item \textbf{Minimum Value (-1):} Reflects poor engagement, occurring when views and likes are at their lowest (both 0) and sentiment is entirely negative (\( P_v = -1 \)).
\end{itemize}

This range helps distinguish high engagement from low engagement across videos and courses, allowing content creators and educators to identify areas for improvement.

\subsubsection{Thresholds for Engagement Quality}
\begin{itemize}
    \item \textbf{Good Engagement:} Typically, an \( E_v \) score above 1.5 can indicate strong engagement, as it implies positive sentiment with relatively high views and likes.
    \item \textbf{Moderate Engagement:} A score between 0.5 and 1.5 suggests average engagement, with a balance of sentiment and interaction.
    \item \textbf{Poor Engagement:} Scores below 0.5 indicate low engagement, characterized by either low sentiment, minimal views and likes, or both.
\end{itemize}

\subsubsection{Rationale for LLM-SEM as an Effective Metric}
The LLM-SEM metric is advantageous because it combines quantitative (views, likes) and qualitative (sentiment) data, offering a comprehensive view of student engagement. By using normalized values and sentiment polarity, this metric enables fair comparison across varying levels of interaction, making it a scalable and data-driven tool for evaluating engagement.

\section{Results and Discussion}
\label{sec:Results_Discussion}

To validate the effectiveness of our proposed LLM-SEM framework, we conducted experiments using a dataset of 16,766 human-annotated sentiments. We evaluated three Large Language Models (LLMs) to determine which model provides the best performance for sentiment analysis, which would then be integrated into the LLM-SEM architecture.

The following LLMs were evaluated:
\begin{itemize}
    \item LLama 3.2B
    \item Gemma 9B
    \item RoBERTa (fine-tuned)
\end{itemize}

For each model, we measured performance in terms of accuracy, recall, and F1-score across three sentiment categories: negative, neutral, and positive. The results are summarized in Table~\ref{tab:results}.

\begin{table}[H]
\setlength{\tabcolsep}{0.8pt}
\caption{Sentiment Classification Results for LLama 3.2B, Gemma 9B, and RoBERTa (fine-tuned)}
\label{tab:results}
\centering
\begin{tabular}{lccc}
\toprule
\textbf{Model} & \textbf{Accuracy} & \textbf{Recall} & \textbf{F1-Score} \\ 
\midrule
\textbf{LLama 3.2B}  & 0.55  & 0.50  & 0.41  \\ 
\textbf{Gemma 9B}     & 0.81  & 0.64  & 0.61  \\ 
\textbf{RoBERTa (Fine-tuned)}      & 0.86  & 0.86  & 0.84  \\ 
\bottomrule
\end{tabular}
\end{table}

\subsection{Model Evaluation Results}

From Table~\ref{tab:results}, it is evident that the fine-tuned RoBERTa model achieved the highest overall accuracy and performance metrics, with an accuracy of 0.86 and an F1-score of 0.84. This suggests that fine-tuning significantly improved RoBERTa's ability to classify sentiment, particularly when compared to the other models.

\textbf{Gemma 9B} also performed well, with an accuracy of 0.81 and reasonable recall and F1-scores. It demonstrated strong performance in handling both negative and positive sentiment but faced challenges with neutral sentiment classification, similar to the other models.

\textbf{LLama 3.2B}, while being effective in classifying positive sentiment, had the lowest overall accuracy (0.55) and struggled the most with neutral sentiment classification, which affected its overall performance.

\subsection{Discussion}

All models demonstrated challenges in accurately predicting neutral sentiment, likely due to the inherent difficulty in distinguishing neutral comments from both positive and negative ones, particularly in complex or context-dependent text data. This is a common issue in sentiment analysis where neutral comments tend to be less distinguishable or are misclassified as either positive or negative.

After fine-tuning, the RoBERTa model's performance improved significantly, which emphasizes the importance of fine-tuning in boosting model accuracy and F1-score for sentiment classification tasks. Despite this improvement, even RoBERTa faced difficulty in handling neutral comments effectively, as shown by the relatively lower performance in this category.

\section{Conclusion}
\label{sec:conclusion}
In conclusion, this study introduced a quantitative approach to assessing student engagement on educational YouTube channels using publicly available data and sentiment analysis models. By analyzing data from multiple educational channels and applying multilingual sentiment analysis, we were able to calculate polarity scores for lessons and courses, offering a comprehensive view of student satisfaction. Our method addressed the shortcomings of traditional feedback mechanisms, providing a scalable and data-driven alternative for evaluating content performance. The integration of polarity scores with views and likes resulted in a well-rounded engagement metric, enabling educators and content creators to identify effective content and areas for improvement. The success of multilingual models in handling diverse languages highlights the robustness of this approach, offering actionable insights for optimizing educational content. This method sets the groundwork for further research in expanding engagement metrics and refining sentiment analysis models for more nuanced understanding in online education environments.

\begin{flushleft}

\end{flushleft}
\end{sloppypar}
\end{document}